\documentclass[11pt]{article}

\usepackage[final]{acl}

\usepackage{times}
\usepackage{soul}
\usepackage{latexsym}
\usepackage{tabularx}
\usepackage{algorithm}
\usepackage[noend]{algpseudocode}
\usepackage{amsmath}
\usepackage{makecell}
\usepackage{graphicx}

\usepackage[font=footnotesize,labelfont=bf]{caption}
\usepackage{xcolor}
\usepackage{booktabs}
\usepackage{multirow}
\usepackage{siunitx}
\usepackage{float}

\newcolumntype{Y}{>{\raggedright\arraybackslash}X}
\usepackage{booktabs,array}
\newcolumntype{L}[1]{>{\raggedright\arraybackslash}p{#1}}
\usepackage{enumitem}
\usepackage{amssymb}% http://ctan.org/pkg/amssymb
\usepackage{pifont}% http://ctan.org/pkg/pifont
\newcommand{\cmark}{\ding{51}}%
\newcommand{\xmark}{\ding{55}}%
\newcolumntype{L}[1]{>{\raggedright\arraybackslash}p{#1}}
\newcommand{\excell}[1]{%
  \begin{tabular}[t]{@{}p{0.1em} p{\dimexpr\linewidth-1.8em\relax}@{}} #1 \end{tabular}%
}
\renewcommand{\arraystretch}{1.1} % a touch of row height

\usepackage{listings}
\usepackage[most]{tcolorbox}
\newcolumntype{Y}{>{\raggedright\arraybackslash}X}

\usepackage[T1]{fontenc}
\usepackage[utf8]{inputenc}

\usepackage{microtype}

\usepackage{inconsolata}

\usepackage{graphicx}

\usepackage{todonotes}

\title{Revisiting Semantic Role Labeling:\\ Efficient Structured Inference with Dependency-Informed Analysis}

\author{Sangpil Youm,
Leah Jones, Bonnie J. Dorr\\
        University of Florida, Gainesville, Florida\\
        \texttt{\{youms,leahjones,bonniejdorr\}@ufl.edu}
}

\begin{document}
\maketitle
\vspace*{-0.5in}
\begin{abstract}
Semantic Role Labeling (SRL) provides an explicit representation of predicate-argument structure, capturing linguistically grounded relations such as
\textit{who did what to whom}. While recent NLP progress has been dominated by large language models (LLMs), these systems often rely on implicit semantic representations, often lacking explicit structural constraints and systematic explanatory mechanisms. 
Traditionally, SRL systems have often relied on AllenNLP; however, the framework entered maintenance mode in December 2022, limiting compatibility with evolving encoder architectures and modern inference requirements.
We revisit structured SRL modeling, introducing a modernized encoder-based framework that preserves explicit predicate-argument structure while enabling inference $10\times$ faster. Using BERT-base, the model attains comparable predictive performance, and RoBERTa and DeBERTa further improve F1 performance within the same framework. We adopt a dependency-informed diagnostic methodology to characterize span-level inconsistencies and conduct a representation-level analysis of LLM behavior under dependency-informed structural signals. Results indicate that dependency cues primarily improve structural stability. Finally, we illustrate how the framework’s explicit predicate-argument structure can support multilingual SRL projection as a downstream application.

\end{abstract}

\section{Introduction}

Semantic role labeling (SRL) 
%is a fundamental task in Natural Language Processing (NLP) that 
identifies an explicit representation of predicate-argument structure, capturing linguistically grounded relations, i.e.,
%, specifying %provides information about 
\textit{who did what to whom}. %It offers essential semantic information within sentences by capturing predicate-argument relations. 
For example:
% \begin{center}
% \footnotesize{
% \vspace*{-.15in}
% \hspace*{-.1in}
% \begin{tabular}{p{.01in}p{2.77in}}
% 1.&\mbox{[}\textbf{ARG0} OpenAI] \textit{released} [\textbf{ARG1} a new language model] [\textbf{ARGM-TMP} in March].
% \end{tabular}
% }
% \end{center}
% \begin{center}
% \footnotesize{
% \vspace*{-.02in}
% \hspace*{-.1in}
% \begin{tabular}{p{.01in}p{2.77in}}
% 2.&\mbox{[}\textbf{ARG0} The research team] \textit{evaluated} [\textbf{ARG1} its multilingual SRL performance] [\textbf{ARGM-LOC} on the Universal PropBank dataset].\end{tabular}
% }
% \end{center}
% \begin{center}
% \footnotesize{
% \vspace*{-.02in}
% \hspace*{-.1in}
% \begin{tabular}{p{.01in}p{2.77in}}
% 3.&\mbox{[}\textbf{ARG1} The model] was \textit{fine-tuned} [\textbf{ARG0} by engineers] [\textbf{ARGM-MNR} using dependency constraints].
% % \vspace*{-0.2in}
% \end{tabular}
% }
% \end{center}

\begin{itemize}
\footnotesize
 \vspace*{-0.1in}
\item \mbox{[}\textbf{ARG0} OpenAI] \textit{released} [\textbf{ARG1} a new language model] [\textbf{ARGM-TMP} in March].
 \vspace*{-0.05in}
\item \mbox{[}\textbf{ARG0} The research team] \textit{evaluated} [\textbf{ARG1} its multilingual SRL performance] [\textbf{ARGM-LOC} on the Universal PropBank dataset].
 \vspace*{-0.05in}
\item \mbox{[}\textbf{ARG1} The model] was \textit{fine-tuned} [\textbf{ARG0} by engineers] [\textbf{ARGM-MNR} using dependency constraints].
\vspace*{-0.1in}
\end{itemize}

\begin{figure}[h]
    \centering
    % \vspace*{-.15in}
    \includegraphics[width=\linewidth]{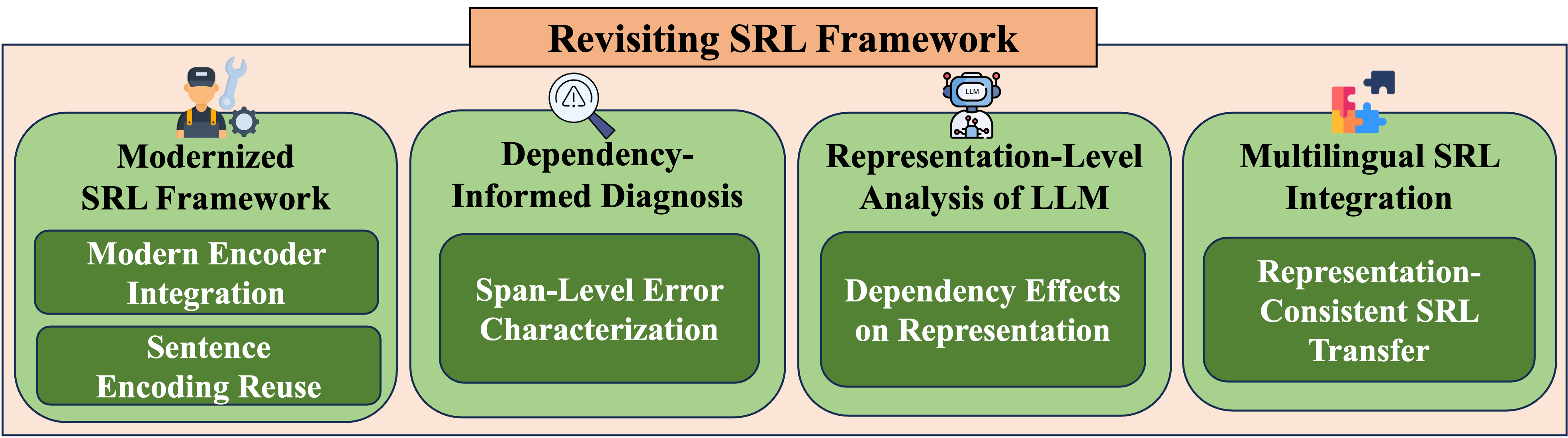}
    % \vspace*{-.15in}
    \caption{Overview of the revisiting SRL framework. Sentence-level encoding reuse enables efficient predicate-conditioned inferences across modern encoder architectures. Dependency-informed diagnosis characterizes span-level errors, and representation-level analysis investigates dependency effects. The framework further supports representation-consistent cross-lingual SRL transfer as a downstream application.
    }%Process of sublingual translations and LLM prompting

    \label{fig:overview}
    \vspace*{-.2in}
\end{figure}
% \vspace*{-.1in}

Following \citet{bonial-etal-2014-propbank}, a \textit{predicate} is assumed to be a verb, noun, adverb, or adjective (e.g., \textit{\textbf{afraid} of bears}), although we focus here on verbs (italicized above). Typical SRL labels are agent (ARG0), theme (ARG1), recipient or secondary argument (ARG2), and modifier %agents (ARG0), themes (ARG1), secondary arguments (ARG2), and modifiers 
(ARGM-TMP/LOC/MNR, etc.).

Due to its linguistic importance, SRL has long been a core NLP benchmark, appearing as shared tasks at CoNLL~\cite{carreras2005conll,hajic2009conll,pradhan2012conll}
% \cite{carreras2005conll,hajic2009conll,pradhan2012conll}. 
This trajectory reflects broader shifts in modeling paradigms, from early statistical systems~\cite{he-etal-2017-deep,li2019dependency} to span-based neural approaches using larger datasets, e.g., PropBank and FrameNet~\cite{conia2022_srl_definition} and transformer encoders~\cite{fei2021end, Zhang2022_SRL_as_Dependency_Parsing}. 
% Modern SRL frameworks leverage contextualized embeddings and syntactic signals to achieve improved structural consistency and predictive performance. 

% Over time, SRL models have evolved from head-word annotations to span-based approaches using BIO tagging and larger datasets, e.g., PropBank and FrameNet~\cite{conia2022_srl_definition}. Early statistical systems~\cite{he-etal-2017-deep,li2019dependency} have given way to %modern 
% transformer-based models~\cite{fei2021end, Zhang2022_SRL_as_Dependency_Parsing}, achieving higher accuracy through contextual embeddings and syntactic integration.
%all of which aim to effectively capture argument structures.

The span-based AllenNLP SRL model~\cite{Gardner2017AllenNLP}, based on~\citet{shi2019simple}, achieves high accuracy and supports
various NLP tasks, including named entity recognition (NER), coreference resolution, and SRL. However, AllenNLP entered maintenance mode in  December 2022, limiting compatibility with evolving encoder architectures and modern inference requirements. 
This obsolescence constrains the deployment of SRL in downstream tasks where predicate-argument features remain valuable---for example, stance detection~\cite{mather-etal-2022-stance}, 
sentiment analysis~\cite{klenner2022semantic},  summarization~\cite{fan-etal-2023-evaluating}, and cross-lingual SRL projection~\cite{youm2024dahrs}. More broadly, they expose a gap between advances in semantic representation learning and the computational efficiency required for large-scale structured inference.

% \begin{figure}[h]
%     \centering
%     \includegraphics[width=\linewidth]{LREC2026 Author's kit/figure/First_figure.png}
    
%     \caption{Overview of the modernized SRL-BERT architecture. The model encodes each sentence once through BERT and reuses cached embeddings for all predicates, removing AllenNLP’s per-predicate instance generation. A dependency-aware module further buckets span-violation errors to support multilingual extension.
%     %Overview of Revisiting SRL. We modernize the SRL-BERT architecture, introduce a syntactic module that buckets span-violation errors, and enable less-constrained multilingual integration to support cross-lingual expansion.
%     }%Process of sublingual translations and LLM prompting
%     % \vspace{-.1in}
%     \label{fig:overview}
%     \vspace*{-.1in}
% \end{figure}

To overcome these integration barriers, we revisit structured SRL modeling by introducing a modernized encoder-based framework (see Figure~\ref{fig:overview}). This design eliminates redundant per-predicate instance generation by caching sentence-level representations and reusing them across predicates, yielding $10\times$ faster inference while preserving predictive performance. Using BERT-base, the model attains comparable performance (86.2\%; AllenNLP: 86.72\%), while RoBERTa~\cite{liu2019roberta} and DeBERTa~\cite{he2020deberta} further improve predictive performance (87.07\%, 87.49\%). Across all evaluated encoder backbones, our framework remains faster than AllenNLP. 

Beyond efficiency and predictive gains, the framework contributes (1) a dependency-informed diagnostic methodology for span boundary inconsistencies and structurally explainable deviations; (2) a representation-level analysis of how dependency-informed structural signals influence LLM SRL predictions, showing improvements in structural stability; and (3) seamless integration with cross-lingual SRL projection pipelines.
%During prediction, we ...
% Figure~\ref{fig:overview}
% \bonnieshort{Hmm, the ref to Figure 1 was missing? I added it. Please edit this figure so it matches the caption.}
% illustrates the architecture and multilingual extension pathway.\footnote{Our Modernized SRL is at:~\url{https://anonymous.4open.science/r/modernized_SRL-33E5}}

% Our main contributions are:

% \begin{center}
% \footnotesize{
% \begin{itemize}
% \item \hl{\textbf{Modernized structured SRL framework}: a dependency-aware, encoder-based architecture enabling sentence-level representation reuse across predicates and compatibility with modern transformer backbones (e.g., BERT, RoBERTa, and DeBERTa).

% \item \textbf{Efficiency gains}: a 10$\times$ faster inference on CoNLL-2012 under the same hardware conditions, with 95\% identical outputs and comparable F1 to the AllenNLP baseline.

% \item \textbf{Dependency-informed diagnostics}: a systematic methodology for characterizing span-level inconsistencies and structurally explainable deviations.

% \item \textbf{Representation-level analysis of LLMs}: an investigation of how dependency-informed structural signals influence LLM SRL predictions, showing improvements primarily in structural stability.

% \item \textbf{Multilingual extensibility}: a seamless integration with cross-lingual SRL projection pipelines.}
% \end{itemize}
% }
% \end{center}

These contributions collectively reposition SRL as a structurally grounded, computationally efficient, and representation-consistent framework, aligning explicit predicate-argument modeling with modern encoder architectures, emerging LLM paradigms, and multilingual SRL transfer.\footnote{Our Modernized SRL is at:~\url{https://anonymous.4open.science/r/modernized_SRL-CF96}}

% modernize SRL for structure (dependency-guided analysis), speed (10$\times$ speedup), \hl{representation alignment (LLM behavior under explicit semantic structure)}, and multilinguality (cross-language projection)\textemdash the tree pillars guiding this work.

% \bonnieshort{Am a little worried that we are light on multilingual in this paper, after making a claim about it in both the abstract and the intro.}

\section{Related Work}
\label{sec:related-work}
SRL research spans two decades, evolving from syntax-driven models to neural span encoders, and more recently LLMs. This evolution reflects broader shifts in NLP modeling paradigms while preserving the core objective of explicit predicate-argument structure representation. To situate our contributions, we review prior work along four dimensions: structured SRL foundations, BERT-based architectures and the AllenNLP framework, LLM-based SRL, and cross-lingual SRL transfer. These threads contextualize our revisiting of structured SRL modeling, emphasizing efficiency, structural consistency, and explanatory adequacy within a dependency-informed, multilingual framework.

\subsection{Structured SRL Foundations }

Early SRL models have emphasized syntactic and lexical cues. Through a series of CoNLL shared tasks~\cite{carreras2005conll,hajic2009conll,pradhan2012conll}
% \cite{carreras2005conll,hajic2009conll,pradhan2012conll}
, these approaches have evolved into neural architectures that incorporate dependency-path features \cite{fitzgerald2015semantic,roth-lapata-2016-neural,he2018syntax}.  Most systems adopt PropBank conventions, assigning ARG0 to the agent, ARG1 to the theme, and ARG2 to the recipient. Over time, these have gradually been replaced by feature engineering with contextual embeddings such as ELMo \cite{peters-etal-2018-deep} and the strong pre-trained baselines like GloVe~\cite{stanovsky-etal-2018-supervised}.

\subsection{BERT-based and AllenNLP SRL}

%The introduction of t
Transformer-based architectures such as BERT \cite{devlin-etal-2019-bert} have allowed for span-level %based formulations using 
BIO tagging as in \citet{conia2022_srl_definition}.  BERT-based %architectures 
models \cite{shi2019simple,fei2021end} perform strongly %performance 
via contextual embeddings and span-level decoding. AllenNLP~\cite{Gardner2017AllenNLP} 
% \cite{Gardner2017AllenNLP} 
generalizes this design for broad community use, integrating SRL, NER, coreference, and parsing. 

Building on these advances, 
%the introduction of transformer-based architectures such as BERT~\cite{devlin-etal-2019-bert} further transforms SRL modeling.  
\citet{shi2019simple} implement a BERT-LSTM model that captures predicates using the [SEP] token, becoming the basis of AllenNLP's SRL implementation. However, post-2022 dependency and package incompatibilities have limited continued integration with modern frameworks
such as spaCy~\cite{spacy} and PyTorch~\cite{paszke2019pytorch}.
% \cite{spacy} and PyTorch~\cite{paszke2019pytorch}. %anguage models 
Our work revisits this encoder-based modeling framework, eliminating redundant per-predicate instance generation, improving inference efficiency, and introducing dependency-informed diagnostics for systematic structural analysis.
% accelerating inference, and introducing dependency-aware diagnostics for systematic error resolution.

%, and we similarly replicate a model using up-to-date libraries and transformer-based architecture, combining BERT embeddings with an LSTM structure. Furthermore, we analyze potential modification points to improve model performance more efficiently.

\subsection{SRL in the LLM Era}

While LLMs excel at sequence modeling, explicit predicate–argument structure remains indispensable for faithfulness, controllability, cross-lingual transfer, and interpretability. SRL %provides these properties by making 
makes semantic relations explicit\textemdash still challenging for LLMs despite their fluency in text generation.
%which remain 

Recent studies show that achieving competitive SRL performance with LLMs requires substantial structural scaffolding, including retrieval mechanisms, frame descriptions, and iterative correction strategies~\citep{Li2025_LLMsDoWell_SRL}. These findings suggest that structured semantic representations do not reliably emerge from generative inference alone.
% explicit structured representations remain central to stable and interpretable semantic modeling, rather than emerging reliably from generative inference alone.  
% (ii) %LLM-based SRL struggles
Prior studies indicate that LLMs struggle to assign consistent semantic roles across domains and languages without an auxiliary mechanism~\citep{wu2022_zeroshot_csrl}, and even advanced models (e.g., GPT-4) attain %achieve 
only partial accuracy % when acting 
as PropBank annotators~\citep{bonn-etal-2024-adjudicating}. Thus, structured role inventories such as PropBank~\cite{Jindal2022_UP2} and Universal PropBank (UP)~\cite{Pradhan2022_PropBank_Comes_of_Age} remain indispensable intermediate representations for reasoning and semantic transfer.
% \citep{Pradhan2022_PropBank_Comes_of_Age}.

Comparative analyses show that models preserving predicate-argument boundaries offer greater interpretability and compositional generalization than purely end-to-end LLMs \citep{Zhang2022_SRL_as_Dependency_Parsing, Shi2020_SRL_as_Syntactic_Dependency_Parsing}.
The recent survey by~\citet{Chen2025_SRL_Survey}  notes renewed interest in structured semantics as a complement to foundation models. SRL thus bridges symbolic semantics and neural contextualization, providing interpretable scaffolding for faithful semantic reasoning.

Building on this perspective, we revisit structured SRL modeling, introducing a modernized encoder-based framework that achieves comparable predictive performance with 10$\times$ faster inference using BERT-base, while RoBERTa and DeBERTa provide further gains. The framework integrates dependency-informed diagnostics for systematic structural analysis and representation-level evaluation of LLM behavior under explicit structural signals.

% \hl{(\texttt{GPT-OSS-120b}. 
% Together, these advances position structured SRL modeling as a lightweight and transparent complement to LLM-based systems, retaining explicit predicate-argument boundaries that large models often obscure.}
% and forming the basis for cross-lingual extensions, and an interpretable complement to LLM-based systems.

% preserving explicit predicate-argument structure, while enabling efficient large-scale semantic analysis.

\subsection{Cross-lingual SRL Transfer}

Cross-lingual SRL transfer is now a central research focus, as annotated data remain scarce for most languages.  
However, such transfer inherently carries over annotation error and structural inconsistencies from English data to the target language. Our approach addresses this problem by detecting structural errors in English source sentences using a dependency-aware error detection method, improving the quality of cross-lingual SRL projection in downstream settings. 

% Building on the structural transparency and dependency alignment of our model, we extend SRL representations to multilingual contexts through projection and role alignment. 

Prior approaches such as divergence-aware SRL projection \cite{youm2024dahrs} and zero-shot conversational SRL \cite{wu2022_zeroshot_csrl} highlight the persistent challenge of maintaining consistent argument structures across typologically diverse corpora. Resources such as Universal PropBank 2.0
% ~\cite{Jindal2022_UP2} 
% \cite{Jindal2022_UP2} 
and PropBank's latest multilingual frames \cite{Pradhan2022_PropBank_Comes_of_Age} now facilitate unified role mappings. These resources support and align with our model's dependency-aware design for cross-lingual expansion.

\section{Structured SRL Framework}
\label{sec:BERT-SRL}

We fine-tune and evaluate our framework on CoNLL-2012. This section describes the dataset and the processing pipeline.
% our model. Below, we describe the dataset and preprocessing pipeline that modernizes the 
%model architectures that distinguish it from the 
% AllenNLP implementation into a more efficient design.

\subsection{Data}

We adopt the English portion of the CoNLL-2012 Shared Task dataset~\cite{pradhan2012conll} derived from OntoNotes 5.0~\cite{weischedel2013ontonotes}, using the official train, development, and test splits for benchmark consistency. The dataset provides span-based semantic role annotations grounded in PropBank, enabling systematic investigation of predicate-argument boundary prediction and structural consistency.\footnote{Dataset creation details are provided in Appendix~\ref{sec:app_dataset}.} 

CoNLL-2012 is selected due to its span-level formulation of semantic roles, which aligns with our framework's emphasis on structural adequacy, argument-boundary stability, and dependency-informed diagnostics. This contrasts with token-level formulations such as CoNLL-2009, which target a different granularity of SRL.
% are less suitable for analyzing span-level structural behavior central to this study.
% adopt a token-level labeling paradign, making them less suitable for analyzing span-level structural behavior central to this study.

\subsection{Model}
% \bonnie{I found this section exceedingly rough and didn't know how to fix it.}
To construct the encoder-based structured SRL model, we condition the input on the predicate by appending a dedicated predicate span to the sentence representation:
\vspace*{-.1in}

\begin{quote}
\footnotesize{}
\texttt{[CLS]|sent|[SEP]|pred|[SEP]}
% \vspace*{-.1in}    
\end{quote}
\vspace*{-.1in}
Here \texttt{sent} and \texttt{pred} denote the sentence and predicate tokens, respectively. We follow AllenNLP training paradigm to ensure comparability, but diverge at inference time by eliminating redundant per-predicate sentence encoding. AllenNLP constructs a separate instances for each predicate, resulting in repeated sentence tokenization and encoder recomputation. In contrast, our framework performs a single sentence-level tokenization and encoder pass and reuses them across all predicates in the sentence.
%runs 
% per-predicate instances, whereas ours tokenizes each sentence once and reuses the cached tokenization across predicates.

For comparison with our approach, we present Algorithm~\ref{algo:allennlp} to illustrate the code underlying AllenNLP's per-predicate encoding system.\footnote{AllenNLP's per-predicate encoding approach is openly available: \url{https://docs.allennlp.org/models/main/models/structured_prediction/predictors/srl/}.}
%SRL inference steps in AllenNLP below:
% \vspace*{-.01in}
\begin{algorithm}[htbp]
\scriptsize
\captionsetup{font=footnotesize}
\caption{Original AllenNLP: Per-Predicate Encoding}
\label{algo:allennlp}
\begin{algorithmic}[1]

\Function{TextToInstance}{$tokens,labels$}
  \State $sent \gets \textsc{HFEncode}(tokens)$
  \State $pred \gets \textsc{HFEncode}(\textsc{Select}(tokens,\ labels{=}1))$
  \State $ids \gets [\texttt{CLS}]\texttt{|sent|}[\texttt{SEP}]\texttt{|pred|}[\texttt{SEP}]$
  \State $tt \gets [0]^{1+\texttt{|sent|}+1} \|[1]^{\texttt{|pred|}+1};\quad am \gets [1]^{|ids|}$
  \State \Return $\textsc{Instance}(\textsc{MakeFields}(ids,tt,am,labels,tokens))$
\EndFunction
\Function{PredictSRL}{$model, tokenizer, inputs, p\_idxs$}
  \State $ INSTS \gets [\,];\ P \gets [\,]$
  % \For{$j=1..B$}
  %   \State $sent \gets inputs[j].sentence$
  %   \State $tokens \gets \textsc{Tokenize}(sent)$
  %   \State $v\_idxs \gets \textsc{FindVerbs}(tokens); \textsc{Push}(VPER,\ v\_idxs)$
  \For{$p \in p\_idxs$}
    \State $labels \gets \textsc{OneHot}(|inputs|,p)$
    \State $inst \gets \Call{TextToInstance}{inputs,labels}$
    \State $\textsc{Push}(INSTS,\ inst)$; $\textsc{Push}(P,\ v\_idxs)$
  \EndFor
  % \EndFor
  % \State $BATCHES \gets \textsc{GroupByCount}(INSTS,B)$
  \State $OUT \gets\textsc{ForwardInstances}(model,INSTS)$
  \State \Return \Call{PackBySentence}{$OUT,\ P$}
\EndFunction
% \vspace*{-.17in}
\end{algorithmic}
% \vspace*{-.17in}
\end{algorithm}
% \vspace*{-.1in}

\begin{center}
\footnotesize{
\vspace*{-.2in}
\hspace*{-.1in}
\begin{tabular}{p{.01in}p{2.77in}}
{\tiny$\bullet$}&\textbf{Line 1:} 
  Define \texttt{TextToInstance} — a helper function responsible for constructing a per-predicate instance for semantic role labeling (SRL).
\end{tabular}
}
\end{center}
\begin{center}
\footnotesize{
\vspace*{-.05in}
\hspace*{-.1in}
\begin{tabular}{p{.01in}p{2.77in}}
{\tiny$\bullet$}&\textbf{Line 2--3:} 
  % \texttt{sent}: 
  Encode the input tokens into subword IDs (\texttt{sent}) using the Hugging Face tokenizer (\texttt{HFEncode}). Select and encode the predicate token (\texttt{pred}).
\end{tabular}
}
\end{center}
\begin{center}
\footnotesize{
\vspace*{-.05in}
\hspace*{-.1in}
\begin{tabular}{p{.01in}p{2.77in}}
{\tiny$\bullet$}&\textbf{Lines 4--5:} 
  Construct the model input sequence:
  concatenate predicate with the sentence, \texttt{[CLS]|sent|[SEP]|pred|[SEP]};
  assign segment IDs (\texttt{tt}) to distinguish sentence and predicate parts (0 for \texttt{sent}, 1 for \texttt{pred});
  create the attention mask (\texttt{am}) to indicate valid tokens.
\end{tabular}
}
\end{center}
\begin{center}
\footnotesize{
\vspace*{-.05in}
\hspace*{-.1in}
\begin{tabular}{p{.01in}p{2.77in}}
{\tiny$\bullet$}&\textbf{Line 6:} 
  Return a structured \texttt{INSTANCE} object containing the constructed fields (\texttt{ids}, \texttt{tt}, \texttt{am}, \texttt{labels}, and \texttt{tokens}) for downstream modeling, and end of the function.
\end{tabular}
}
\end{center}
\begin{center}
\footnotesize{
\vspace*{-.05in}
\hspace*{-.1in}
\begin{tabular}{p{.01in}p{2.77in}}
{\tiny$\bullet$}&\textbf{Line 7:} 
  Define \texttt{PredictSRL(model, tokenizer, inputs, p\_idxs)} — the main SRL inference routine, which processes input sentences and predicate indices.
\end{tabular}
}
\end{center}
\begin{center}
\footnotesize{
\vspace*{-.05in}
\hspace*{-.1in}
\begin{tabular}{p{.01in}p{2.77in}}
{\tiny$\bullet$}&\textbf{Line 8:} 
  Initialize an empty list of per-predicate instances (\texttt{INSTS}) and a mapping of predicates to their source sentences (\texttt{P}).
  % \footnotemark
\end{tabular}
}
\end{center}
% \footnotetext{Although we consider predicates to extend beyond verbs, the original AllenNLP only maps verbs to predicates.}
\begin{center}
\footnotesize{
\vspace*{-.05in}
\hspace*{-.1in}
\begin{tabular}{p{.01in}p{2.77in}}
{\tiny$\bullet$}&\textbf{Lines 9--12:} 
  Iterate over all predicate indices. For each predicate position \texttt{p}:
  create a one-hot label vector (\texttt{labels}); build a per-predicate instance using \texttt{TextToInstance} function (line 1-7); append the instance to \texttt{INSTS}.
\end{tabular}
}
\end{center}
\begin{center}
\footnotesize{
\vspace*{-.05in}
\hspace*{-.1in}
\begin{tabular}{p{.01in}p{2.77in}}
{\tiny$\bullet$}&\textbf{Line 13:} 
  Perform a forward pass over the model with all instances (\texttt{INSTS}) to generate per-predicate predictions (\texttt{OUT}).
\end{tabular}
}
\end{center}
\begin{center}
\footnotesize{
\vspace*{-.05in}
\hspace*{-.1in}
\begin{tabular}{p{.01in}p{2.77in}}
{\tiny$\bullet$}&\textbf{Line 14:} 
  Regroup predictions by their original sentences using \texttt{PackBySentence(OUT, P)} and return the final output: a list of SRL-labeled, BIO-marked sentences.
\end{tabular}
}
\end{center}

% In Algorithm~\ref{algo:our_model}, we encode the sentence a single time, caching the sentence ($sent$), the first wordpieces index ($f\_wp\_idxs$), and the word count ($n\_words$) via \textsc{EncodeSentenceOnce} function (see line 9). For each verb, we tokenize only that word ($p\_wp$) and build input ids ($ids$) in the same format as in Algorithm~\ref{algo:allennlp}, with $tt$ and $am$. We stack all per-verb inputs ($IDS/TT/AM$) and pass them to the model together, along with the auxiliary tensor ($AUX = \{P, WF, n\_words\}$). The model returns logits ($LOGITS$), which we decode with id-to-label ($i2l$) to obtain $TAGS$. This avoids per-verb sentence re-tokenization and instance plumbing, yielding faster prediction with equivalent inputs.

% Our modernized SRL model follows 
Algorithm~\ref{algo:our_model} presents our modernized inference design, which decouples sentence encoding from predicate-conditioned prediction. A single sentence-level encoding pass is performed, ensuring consistent token indices across all predicate-conditioned structures within a sentence. The cached sentence representation is reused to construct predicate-conditioned inputs. Importantly, each predicate still receives a predicate-conditioned contextualized encoder representation, preserving predicate-specific modeling capacity while eliminating redundant preprocessing.

% predicates  improving inference efficiency by caching 
% contextual embeddings.
% sentence-level tokenization
% \footnote{Our 
% % Sentence 
% % Tokenization approach 
% Modernized SRL is available at this anonymous link:~\url{https://anonymous.4open.science/r/modernized_SRL-33E5}} 
% and reusing it across predicates. Unlike AllenNLP’s per-predicate
% re-encoding (of both sentence and predicate), our model runs a single tokenization pass per sentence and constructs all predicate inputs from the cached sentence representation, reducing redundant preprocessing while maintaining contextual accuracy. Each predicate still triggers a contextual encoding through BERT, ensuring predicate-specific representations.
% with each predicate invoking a lightweight classification step.

\begin{algorithm}[]
\scriptsize
% \captionsetup{font=footnotesize}
\caption{Our Approach: Modernized SRL}
\label{algo:our_model}
\begin{algorithmic}[1]

\Function{EncodeSentenceOnce}{$inputs, tokenizer$}
  \State $wp \gets tokenizer(inputs,\ split=T,\ specials=F)$
  \State $sent \gets \textsc{Flatten}(wp.ids)$
  \State $f\_wp\_idxs \gets \textsc{FirstPosPerWord}(wp.word\_ids(),\ +1)$
  % \State $n\_words \gets |inputs|$
  % \State \Return $(sent,\ f\_wp\_idxs,\ n\_words)$
  \State \Return $(sent,\ f\_wp\_idxs)$
\EndFunction

\Function{PredictSRL}
{{\mbox{$model, tokenizer, inputs, p\_idxs$}}}

  % \State $
  %  \begin{aligned}[t]
  %   (sent,\ f\_wp\_idxs,\ n\_words) &\gets \Call{EncodeSentenceOnce}{}\\[-2pt]
  %                                    &\quad (inputs,\ tokenizer)
  %   \end{aligned}
  %   $

  \State $
   \begin{aligned}[t]
    (sent,\ f\_wp\_idxs) &\gets \Call{EncodeSentenceOnce}{}\\[-2pt]
                                     &\quad (inputs,\ tokenizer)
    \end{aligned}
    $
     
 % \State $IDS \gets [\,],\ TT \gets [\,],\ AM \gets [\,],\ P \gets [\,],\ WF \gets [\,]$
  \For{$p \in p\_idxs$}
    \State $IDS,TT,WF,AM,P \gets [\,]$
    % \State $
    % \begin{aligned}[t]
    % pred \gets{}&\ \textsc{Flatten}(\Call{tokenizer}{[\,inputs[p]\,],\ split=T, \\
    % & specials=F}.ids)
    % \end{aligned}
    \State $
    \begin{aligned}[t]
    pred \gets{}&\ \textsc{Flatten}(\Call{tokenizer}{inputs[p]})
    \end{aligned}
    $
    \State$
    \begin{aligned}[t]
    ids &\gets [\texttt{CLS}] \texttt{|sent|} [\texttt{SEP}] 
                        %\\
                        %&\ \ 
                        \texttt{|pred|} [\texttt{SEP}]
    \end{aligned}$
    \State $tt \gets [0]^{\,1+\texttt{|sent|}+1} \,\|\, [1]^{\,\texttt{|pred|}+1}$, $am  \gets [1]^{\,\lvert ids\rvert}$
    \State $\textsc{Push}(IDS, ids);$ 
    % \State 
    $\textsc{Push}(TT, tt);\ \textsc{Push}(AM, am);$
    \State $\textsc{Push}(\textit{P}, p); 
    \textsc{Push}(WF, f\_wp\_idx)$
    
  \EndFor
  \State $INPUTS \gets \textsc{PadAndStack}(IDS,\ TT,\ AM)$
  % \State $AUX \gets \textsc{BuildAux}(P,\ WF,\ n\_words)$
  % \State $LOGITS \gets \textsc{Forward}(model,\ INPUTS,\ AUX)$
    %  \begin{aligned}[t]
    % pred \gets{}&\ \textsc{Flatten}(\Call{tokenizer}{[\,inputs[p]\,],\ split=T, \\
    % & specials=F}.ids)
    % \end{aligned}
  % \begin{aligned}[t]  
  \State $OUT \gets \textsc{Forward}(model,\ INPUTS,\
                                            P,\ WF)$
  % \end{aligned}
  % \State $TAGS \gets \textsc{Map}(\textsc{ArgmaxLast}(LOGITS),\ i2l)$
  \State \Return \Call{PackBySentence}{$OUT,P$}
\EndFunction

\end{algorithmic}
% \vspace*{-.15in}
\end{algorithm}

\begin{center}
\footnotesize{
% \vspace*{-.1in}
\hspace*{-.1in}
\begin{tabular}{p{.01in}p{2.77in}}
{\tiny$\bullet$}&\textbf{Line 1:} 
  Define \texttt{EncodeSentenceOnce}\textemdash a helper function that performs sentence-level encoding once, avoding redundant recomputation across predicates.
\end{tabular}
}
\end{center}
\begin{center}
\footnotesize{
\vspace*{-.05in}
\hspace*{-.1in}
\begin{tabular}{p{.01in}p{2.77in}}
{\tiny$\bullet$}&\textbf{Lines 2--5:} 
  Perform sentence-level encoding and preprocessing: tokenize into wordpieces, flatten token IDs, and record first wordpiece indices. 
  % and total word count.
  Then, return preprocessed sentence representation (\texttt{sent}) and wordpiece index mapping (\texttt{f\_wp\_idxs}). 
 \end{tabular}
}
\end{center}
% \begin{center}
% \footnotesize{
% \vspace*{-.05in}
% \hspace*{-.1in}
% \begin{tabular}{p{.01in}p{2.77in}}
% {\tiny$\bullet$}& 
%   %\begin{itemize}[leftmargin=2em]
%    %   \item Tokenize sentence into wordpiece tokens.
%    %   \item Flatten token IDs into a single sequence.
%    %   \item Identify first wordpiece index for each word.
%    %   \item Record total number of words.
%   %\end{itemize}
% \textbf{Line 6:} 
%   Return preprocessed sentence representation (\texttt{sent}), wordpiece index mapping (\texttt{f\_wp\_idxs}), and word count (\texttt{n\_words}).
% \end{tabular}
% }
% \end{center}
\begin{center}
\footnotesize{
\vspace*{-.05in}
\hspace*{-.1in}
\begin{tabular}{p{.01in}p{2.77in}}
{\tiny$\bullet$}&\textbf{Line 6:} 
  Define main SRL inference function\textemdash \texttt{(PredictSRL(model, tokenizer, inputs, v\_idxs)}.
\end{tabular}
}
\end{center}
\begin{center}
\footnotesize{
\vspace*{-.1in}
\hspace*{-.1in}
\begin{tabular}{p{.01in}p{2.77in}}
{\tiny$\bullet$}&\textbf{Line 7:} 
  Encode the sentence once using \texttt{EncodeSentenceOnce} (line 1-7) to avoid reundant sentence-level encoding and improve efficiency.
\end{tabular}
}
\end{center}
\begin{center}
\footnotesize{
\vspace*{-.05in}
\hspace*{-.1in}
\begin{tabular}{p{.01in}p{2.77in}}
{\tiny$\bullet$}&\textbf{Lines 8--14:} 
  Iterate over each predicate and build model inputs: initialize empty containers, \texttt{IDS}, \texttt{TT}, \texttt{WF}, \texttt{AM}, and \texttt{P}; tokenize the predicate span; %\texttt{pred} $\leftarrow$ \texttt{FLATTEN(tokenizer([inputs[p]], split=T, specials=F).ids)}:
  and construct full input by concatenating the reused sentence token IDs with predicate token ID, \texttt{[CLS]|sent|[SEP]|pred|[SEP]};
  build segment IDs (0 for \texttt{sent}, 1 for \texttt{pred}); create the attention mask (\texttt{am}) to indicate valid tokens. %\texttt{am = [1]$^{|ids|}$}; Store the constructed components with \texttt{Push} operations into their respective lists.
\end{tabular}
}
\end{center}
\begin{center}
\footnotesize{
\vspace*{-.05in}
\hspace*{-.1in}
\begin{tabular}{p{.01in}p{2.77in}}
{\tiny$\bullet$}&\textbf{Line 15:} 
  Combine the per-predicate input tensors into a batch with \texttt{PadAndStack(IDS, TT, AM)} for model inference.
\end{tabular}
}
\end{center}
% \begin{center}
% \footnotesize{
% \vspace*{-.05in}
% \hspace*{-.1in}
% \begin{tabular}{p{.01in}p{2.77in}}
% {\tiny$\bullet$}&\textbf{Line 17:} 
%   Build auxiliary features (\texttt{AUX}) from predicate positions (\texttt{P}), wordpiece mapping (\texttt{WF}), and total word count (\texttt{n\_words}).
% \end{tabular}
% }
% \end{center}
\begin{center}
\footnotesize{
\vspace*{-.05in}
\hspace*{-.1in}
\begin{tabular}{p{.01in}p{2.77in}}
{\tiny$\bullet$}&\textbf{Line 16:} 
  Perform the forward pass: \texttt{OUT} $\leftarrow$ \texttt{Forward(model, INPUTS, P, WF)}, producing contextualized token representations that are mapped to predicate-specific logits.
\end{tabular}
}
\end{center}
% \begin{center}
% \footnotesize{
% \vspace*{-.05in}
% \hspace*{-.1in}
% \begin{tabular}{p{.01in}p{2.77in}}
% {\tiny$\bullet$}&\textbf{Line 19:} 
%   Map the highest-scoring logits to semantic roles 
%   % \texttt{TAGS} $\leftarrow$ \texttt{Map(ArgmaxLast(LOGITS), i2l)}.
% \end{tabular}
% }
% \end{center}
\begin{center}
\footnotesize{
\vspace*{-.05in}
\hspace*{-.1in}
\begin{tabular}{p{.01in}p{2.77in}}
{\tiny$\bullet$}&\textbf{Line 17:} 
  % Return the final outputs: predicate positions (\texttt{P}), predicted argument tags (\texttt{TAGS}), and model logits (\texttt{LOGITS}).
  Regroup predictions by their original sentences using \texttt{PackBySentence(OUT, P)} and return the final output: a list of SRL-labeled, BIO-marked sentences.
\end{tabular}
}
\end{center}
Although sentence-level representations are shared, predicate-specific distinctions are introduced via the predicate span and segment embeddings, ensuring that argument predictions remain predicate-conditioned rather than sentence-global. We next evaluate the predictive performance and efficiency of our model on benchmark data.

\section{Experiments and Results}

% In this section, we describe the experimental setup of our model.
We follow the AllenNLP SRL configuration 
%approach used in the AllenNLP implementation of SRL, which is 
based on the SRL-BERT model %developed
by~\citet{shi2019simple}. 
%Our experimental configuration is aligned with the settings presented in that work. Specifically, we set the 
The LSTM and MLP hidden dimensions are 768 and 300, respectively, with
%and use 
a predicate indicator embedding size of 10. 
%The learning rate is configured to 
Using a BERT-base-cased encoder and a learning rate of $5 \times 10^{-5}$, we conduct all  
%and the experiments are conducted using the BERT-base-cased encoder. We run %our 
experiments on 
%three 
AMD EPYC 9655P (3 CPUs) and NVIDIA DGX B200 machines (1 GPU). 
% \bonnieshort{I edited here, but if this is supposed to be 3 nodes, change to three DGX B200 nodes ?}

On the test set (11,289 sentences), our model runs 10$\times$ faster (1.32 min vs.\ 13.18 min) than the AllenNLP baseline, with nearly identical accuracy. These sentences potentially contain more than one clause and 
%on 
average 26 tokens, totaling 32,617 clause-level predictions and 877,873 tokens. At the phrase level, we achieve a comparable F1: 86.15\% (ours) vs.\ 86.72\% (AllenNLP's), with Precision/Recall scores of 85.98/86.32\%  (ours) vs. 86.81/86.63\% (AllenNLP). Table~\ref{tab:results} summarizes these results.
% \begin{table}[h]
% \centering
% \small
% \begin{tabular}{lrrr}
% \toprule
% Metric & \multicolumn{1}{c}{Our Model (BERT)} & \multicolumn{1}{c}{Our Model (RoBERTa)} & \multicolumn{1}{c}{AllenNLP} \\
% \midrule
% P (\%)  & 85.98  & \textbf{87.03} & 86.81 \\
% R (\%)  & 86.32  & \textbf{87.52}& 86.63 \\
% F1 (\%)  & 86.15  & \textbf{87.27} & 86.72 \\
% Time (m) & \textbf{1.36}  & 13.18 \\
% \bottomrule
% \end{tabular}
% \caption{Comparison of our modernized BERT-based SRL model and AllenNLP on the OntoNotes 5.0 test set, showing comparable accuracy with a 10$\times$ speedup.}
% \label{tab:results}
% \end{table}

\begin{table}[h]
\centering
\small
\resizebox{\columnwidth}{!}{
\begin{tabular}{lcccc}
\toprule
Metric & \makecell{Our Model \\ (BERT)} & \makecell{Our Model \\(RoBERTa)} & \makecell{Our Model \\(DeBERTa)}  & AllenNLP \\
\midrule
P (\%)   & 85.98 & 86.19 & \textbf{86.78} & 86.81 \\
R (\%)   & 86.32 & 87.97 & \textbf{88.22} & 86.63 \\
F1 (\%)  & 86.15 & 87.07 & \textbf{87.49} & 86.72 \\
Time (m) & \textbf{1.36} & 5.36  & 7.57 & 13.18 \\
\bottomrule
\end{tabular}
}
\caption{Comparison of our modernized BERT-based SRL model and AllenNLP on the OntoNotes 5.0 test set, showing comparable accuracy with a 10$\times$ speedup.}
\label{tab:results}
\end{table}
\vspace*{-.1in}

We use BERT-based model as the primary comparison model against AllenNLP because BERT represents the community-standard baseline transformer encoder for SRL and offers the most efficient inference speed among the evaluated encoders. RoBERTa and DeBERTa are included to assess performance sensitivity to backbone choice. Across encoder variants, predictions exhibit high agreement with AllenNLP outputs (BERT: 95.2\%, RoBERTa: 97.97\%, DeBERTa: 94.99\%)

Of the 877,873 tokens, 841,002 tokens (95.2\%) match AllenNLP outputs on the same unseen test set, confirming replication fidelity. The remaining 4.8\% (42,174 tokens) represent token-level mismatches analyzed next.
% \bonnieshort{Add 100th place decimal for all results in this paragraph and in the table.}

Table~\ref{tab:agreement} presents a breakdown of the distribution of these predictions relative to the OntoNotes 5.0 ground truth, distinguishing between agreement and disagreement cases, as well as whether each system is correct. Among agreement cases\textemdash where both systems produce identical outputs\textemdash88.49\% are correct (accounting for 84.24\% of the total), and 11.51\% are incorrect (10.96\% of the total). In the disagreement cases, 43.96\% of AllenNLP's predictions are correct (2.11\% of the total), and 29.29\% of our system are correct (1.41\% of the total). 
% Cases where neither system is correct account for the remaining 
The remaining 26.75\% (1.28\% of the total) are cases where neither system is correct. 
% Overall, our model achieves an accuracy of 85.56\%, compared to 86.35\% for AllenNLP.

% \begin{table}[t]
% \centering
% \begin{tabular}{l l cc}
% \toprule
% Partition & Subcase & \% within partition & \% of total \\
% \midrule
% \multirow{2}{*}{Agreement (95.20\%)} 
% & Both correct      & 88.49 & 84.24 \\
% & Both wrong        & 11.51 & 10.96 \\[2pt]
% \multirow{3}{*}{Disagreement (4.80\%)} 
% & Allen correct     & 43.96 & 2.11 \\
% & Ours correct      & 29.29 & 1.41 \\
% & Neither correct   & 26.75 & 1.28 \\
% \bottomrule
% \end{tabular}
% \caption{Breakdown of agreement vs.\ disagreement with correctness relative to the OntoNotes 5.0 ground truth.}
% \label{tab:agreement}
% \end{table}

\begingroup
\setlength{\tabcolsep}{4pt}     % default ~6pt; tighter
\renewcommand{\arraystretch}{1.05}

\begin{table}[h!]
\centering
\scriptsize{}
\begin{tabularx}{\linewidth}{@{}Y Y r r@{}}
\toprule
\textbf{Partition} &
\textbf{Subcase} &
\makecell{\textbf{\% within}\\\textbf{partition}} &
\makecell{\textbf{\% of}\\\textbf{total}} \\
\midrule
\multirow{2}{*}{\excell{Agreement (95.20\%) }} &
Both correct            & 88.49\% & 84.24\% \\
& Both wrong             & 11.51\% & 10.96\% \\

\midrule
\multirow{3}{*}{\excell{Disagreement (4.80\%) }} &
Allen correct        & 43.96\% & 2.11\% \\
& Ours correct          & 29.29\% & 1.41\% \\
&Neither correct         &26.75\% & 1.28\% \\
\bottomrule
\end{tabularx}
\caption{Prediction outcomes relative to the ground truth are divided into \textit{agreement} (835,699 tokens), where both systems produce identical predictions, and \textit{disagreement} (42,174 tokens), where their outputs diverge. Within the agreement set (95.2\% of the total), both systems are correct for 734,415 tokens (88.49\% of the within agreement). Within the disagreement set, AllenNLP is correct for 18,539 tokens (43.96\%), and our system for 12.352 tokens (29.29\%).}
\label{tab:agreement}
\end{table}
% \vspace*{-.15in}
\endgroup
\vspace*{-.2in}

\section{Analysis and Remediation}

Although overall performance is strong, understanding residual errors is essential. 
To analyze mismatches, we employ a \textit{dependency-aware} analyzer that automatically merges repeated spans
when possible, and flags the rest for human review. This process resolves a measurable subset of mismatches and systematically categorizes the rest.
%while providing a systematic framework for categorizing the rest. 
Fine-grained character-, token-, and span-level comparisons make manual review impractical. Our analyzer distinguishes between automatic and human-resolvable errors, thereby enhancing the structural interpretability of model behavior. 

Before detailing the dependency-aware error detection framework, we first examine which semantic role categories are most frequently missed in the 5\% of predictions where our system and AllenNLP diverge. This analysis lays the foundation for characterizing the sources of residual errors.
% \vspace*{-.08in}
\subsection{Analysis of Missing Roles in Ours and AllenNLP}
% \vspace*{-.08in}
We investigate missed arguments via one-to-one comparisons between our system and AllenNLP outputs, evaluated against the OntoNotes 5.0 ground truth. 
Relative to our system, \texttt{ARGM} accounts for 48.05\% of differences across argument categories. Core roles ARG0-ARG2 together account for 48.77\% of mismatches, with \texttt{ARG1} and ARG2 most prominent. Among \texttt{ARGMs}, \texttt{ARGM-ADV} (adverbial modifier) represents the largest share of discrepancies (28.33\%), followed by \texttt{ARGM-MNR} (11.97\%) and \texttt{ARGM-TMP}, with \texttt{ARGM-LOC} forming the next substantial portion.\footnote{Full diagnostic distributions are provided in Appendix~\ref{sec:missing_roles_our}.}

We next present role-wise differences relative to AllenNLP, evaluated against the ground truth. The largest share of missing labels is \texttt{ARGM} (48.10\%). The core roles \texttt{ARG0}, \texttt{ARG1}, and \texttt{ARG2} together represent 48.42\% of differences. Within \texttt{ARGMs}, a small subset of adjunct types dominates missing labels: \texttt{ARGM-ADV} (25.72\%), \texttt{ARGM-PRD} (13.13\%), followed by \texttt{ARGM-MNR}, \texttt{ARGM-TMP}, and \texttt{ARM-LOC}\textemdash accounting for over 75\% of ARGM-related omissions.\footnote{Full diagnostic distributions are provided in Appendix~\ref{sec:missing_roles_allen}.}

% referring to purpose clauses for the motivation for some action. Such clauses are typically marked by phrases like ``so that''. The next largest contributions are \texttt{ARGM-MNR}, \texttt{ARGM-TMP}, and \texttt{ARM-LOC}. Taken together, these five adjunct types exceed 75\% of the missing labels.\footnote{Full diagnostic distributions are provided in Appendix~\ref{sec:appendixA}.} 

While this analysis identifies dominant omission patterns, it does not explain the structural causes. The next section, therefore, introduces a dependency-aware diagnostic framework to examine span-level sources of inconsistency.

% error analysis that examines structural sources of span-level mistakes.

\subsection{Dependency-Aware Error Analysis}

\begin{table*}[htbp]
\scriptsize
\centering
\small
%            Subtype      Definition      Example (wider)
\begin{tabular}{@{}L{0.16\linewidth} L{0.2\linewidth} L{0.60\linewidth}@{}}
\toprule
\textbf{Subtype} & \textbf{Definition} & \textbf{Example} \\
\midrule

\excell{same\_head} &
Both spans attach to the same syntactic head. &
% \excell{\emph{\cmark\ Ours: [ARG2: just brain dead]}\\
%              \emph{\xmark\ Allen: [ARG2: just] [ARG2: brain dead]}\\
%              \emph{\cmark\ Ours: [ARG2: just brain dead]}\\
%              \emph{\xmark\ Allen: [ARG2: just] [ARG2: brain dead]}
%              } \\[2pt]

\excell{
  \cmark&\emph{Ours: [ARG2: just brain dead]}\\
  \xmark&\emph{Allen: [ARG2: just] [ARG2: brain dead]}\\
  \xmark&\emph{Ours: [ARG1: the various revisions] to [ARG1: trade law]}\\
  \cmark&\emph{Allen: [ARG1: the various revisions to trade law]}
} \\[2pt]
\midrule
\excell{pp\_attach} &
Right (left) span is a PP modifying the left (right) span. &
% \makecell[l]{\emph{\cmark\ Ours: [ARG1: a long section of road]}\\
%             \emph{\xmark\ Allen: [ARG1: a long section] [ARG1: of road]}
%             \emph{\cmark\ Ours: [ARG1: a long section of road]}\\
%             \emph{\xmark\ Allen: [ARG1: a long section] [ARG1: of road]}
%             } \\[2pt]
\excell{
  \cmark&\emph{Ours: [ARG1: a long section of road]}\\
  \xmark&\emph{Allen: [ARG1: a long section] [ARG1: of road]}\\
  \xmark&\emph{Ours:  [ARG1: several kinds] of [ARG1: microbes]}\\
  \cmark&\emph{Allen: [ARG1: several kinds of microbes]}
} \\[2pt]
\midrule
\excell{subtree\_attach} &
One span lies inside the dependency subtree of the other span's head. &
% \makecell[l]{\emph{\cmark\ Ours: [ARGM-TMP: for the fourth time on Friday]}\\
%              \emph{\xmark\ Allen: [ARGM-TMP: for the fourth time] [ARGM-TMP: on Friday]} \\
%              \emph{\xmark\ Ours: [ARGM-TMP: time] in [ARM-TMP: the future]}\\
%              \emph{\cmark\ Allen: [ARGM-TMP: time in the future]}} \\[2pt]

\excell{
  \cmark&\emph{Ours: [ARGM-TMP: for the fourth time on Friday]}\\
  \xmark&\emph{Allen: [ARGM-TMP: for the fourth time] [ARGM-TMP: on Friday]}\\
  \xmark&\emph{Ours: [ARG0: anyone who] finds [ARG0: the pair]}\\
  \cmark&\emph{Allen: [ARG0: anyone who finds the pair]}
} \\[2pt]

% \midrule
% heuristic\_prep(NP) &
% Right span begins with a preposition, forming a lightweight PP. &
% \makecell[l]{\emph{\cmark Allen: [ARGM-TMP: hurry to till]} \\
%             \emph{\xmark Ours: [ARGM-MNR: hurry] [ARGM-MNR: to till]}} \\
% \midrule

% \excell{vp\_subtree\_attach (VP)}&
% One span lies within the other verb’s subtree. &
% % \makecell[l]{\emph{[ARG2: exactly] [ARG2: sure what it is now]}} 

% \excell{
%   \cmark&\emph{Ours: [ARG2: exactly sure what it is now]}\\
%   \xmark&\emph{Allen: [ARG2: exactly] [ARG2: sure what it is now]}\\
%   \xmark&\emph{Ours: [ARG1: know] [ARG1: what this all comes down to] }\\
%   \cmark&\emph{Allen: [ARG1 know what this all comes down to]}
% } \\[2pt]

% \midrule
% \excell{vp\_same\_head (VP)} &
% Both spans are headed by the same verb or auxiliary. &
% % \makecell[l]{\emph{[ARG1 know] [ARG1 what this all comes down to]}} \\[2pt]

% \excell{
%   \cmark&\emph{Ours: [ARGM-TMP: when he was a bird before he was a human]}\\
%   \xmark&\emph{Allen: [ARGM-TMP: when he was a bird] [ARGM-TMP: before he was a human]}\\
%   \xmark&\emph{Ours: [ARG1: know] [ARG1: what this all comes down to]}\\
%   \cmark&\emph{Allen: [ARG1 know what this all comes down to]}
% } \\[2pt]

\bottomrule
\end{tabular}
\caption{Error subtypes, definitions, and illustrative examples. All types reflect \textbf{repeated same arguments} within a sentence. These examples are drawn from 4.8\% of cases where predictions disagree between AllenNLP and our system. In this example, we present a case in which one of the two systems produces the correct prediction.}
\label{tab:error-types-subtypes}
% \vspace*{-.1in}
\end{table*}

% \sangpil[]{Table 3 - add }

Our model predicts span-based semantic roles, 
and some errors arise from BIO tagging mistakes that result in span violations. In particular, we detect cases where the beginning of an argument is incorrectly assigned multiple times to the same semantic role, as illustrated here for \texttt{B-ARG0}. Our analyzer identifies and corrects errors by enforcing proper span boundaries using dependency information. The opening bracket (\texttt{``[''}) marks the start of a span, and all subsequent tokens within that span receive \texttt{I}-tags until the closing bracket (``\texttt{]}'').
\vspace*{-.08in}
\begin{quote}
\footnotesize{
\texttt{[B-ARG0: temperament]}
\texttt{[B-ARG0: through I-ARG0: study I-ARG0: and I-ARG0: practice]}}
\end{quote}
\vspace*{-.08in}
% \bonnieshort{Please identify how often this happens in the 5\% difference that you found, compared to ground truth, for \textbf{each of the two systems} : Allen and Ours.}
Among 4.8\% of divergent predictions between AllenNLP and our system, span violations account for 3.8\% of all spans in all systems' predictions (17,387 spans) and 3.79\% of all spans in the AllenNLP outputs (19,131 spans). In both systems, \texttt{ARGM-ADJ} exhibits the highest number of span violations, representing 16.67\% of such cases in our model and 10.00\% in AllenNLP. 

Because PropBank-style annotation prohibits duplicate argument roles per predicate (e.g., two ARG0 spans for the same verb)~\cite{jurafsky_speech_2025}, repeated spans for the same semantic role are structurally invalid. This constraint\textemdash long established in linguistic theory\textemdash requires that each argument occur only once per predicate. Thus, when our system produces multiple spans for the same role, it merges them into a single continuous argument span to satisfy this constraint. 
%instead should be combined into a single argument:  

% two adjacent spans share the same semantic role, they violate the single-argument constraint that each argument must occur no more than once per predicate. 
% In such cases, the spans together comprise a single argument, 
In such cases, the first token %should carry 
carries the \textit{B} tag, and all subsequent tokens carry the \textit{I} tag, rather than restarting with \textit{B}, as illustrated below:
% As illustrated below, our error analysis focuses on cases in which multiple \texttt{B}-tags are incorrectly assigned to the same role within a single sentence.
% % arguments are assigned to the same role within a single sentence. 
%Here, two separate B-ARG0 tags are incorrectly assigned, even though they should constitute a single continuous argument span. Therefore, they need to be combined into a single span as shown below.
\vspace*{-.08in}
\begin{quote}
\footnotesize{\texttt{[B-ARG0: temperament I-ARG0: through I-ARG0: study I-ARG0: and I-ARG0: practice]}}
\end{quote}
\vspace*{-.08in}
Our error analyzer operates over BIO-tagged token sequences and dependency parsers.\footnote{See Appendix~\ref{appendix:error} for the analyzer pipeline.} It identifies adjacent spans assigned the same semantic role as candidate repeated arguments, whose dependency structure determines classification. Spans attaching to the same syntactic head are labeled as \texttt{same\_head}; cases involving prepositional modification are categorized as \texttt{pp\_attach}; and spans lying within another's dependency subtree are classified as \texttt{subtree\_attach} (see Table~\ref{tab:error-types-subtypes}).  

Errors classified into predefined types are marked ``Fixable''; otherwise, sentences are flagged ``Review Required'', indicating that the error cannot be automatically resolved. Classification outcomes enable structured evaluation of error patterns across systems. We apply dependency-aware error detection to cases where AllenNLP predicts correct tags, but our model does not\textemdash accounting for 2.11\% of tokens within the 4.8\% total disagreement. Over 18,539 such errors, our analyzer identifies 498 tokens (2.68\%) as belonging to fixable categories (\texttt{same\_head}, \texttt{subtree\_attach}, and \texttt{pp\_attach}) that can be automatically resolved. Additionally, 933 tokens (5.03\%) involve repeated role spans that cannot be resolved automatically and are therefore flagged as ``Review Required'' (see Table~\ref{tab:ours_error_bucket_distribution}). 
\begin{table}[htbp]
\scriptsize
\centering
\small
\begin{tabular}{lrr}
\toprule
\textbf{Error bucket} & \textbf{Tokens} & \textbf{\% of 18{,}539} \\
\midrule
NO\_BUCKET         & 17{,}108 & 92.28 \\
OTHER\_REPEAT      &    933 &  5.03 \\
same\_head     &    314 &  1.67 \\
subtree\_attach &     141 &  0.76 \\
% dep\_same\_head    &     69 &  0.37 \\
% vp\_subtree\_attach&     54 &  0.29 \\
pp\_attach         &     43 &  0.23 \\
\bottomrule
\end{tabular}
\caption{Token distribution for cases where AllenNLP is correct on disagreements with our model (18,539 instances). Errors in the five structural categories (\texttt{same\_head}, \texttt{subtree\_attach}, and \texttt{pp\_attach}) are labeled ``Fixable'', while others are flagged as ``Review Required,'' indicating a need for human intervention.}
\label{tab:ours_error_bucket_distribution}
\vspace*{-.2in}
\end{table}

The dependency-aware analyzer provides a systematic framework for characterizing residual inconsistencies. Leveraging structural signals, it identifies potential mislabeling, flags cases needing human judgment, and improves the structural interpretability of SRL predictions.

\begin{table*}[!htpb]

\scriptsize
\centering
% \small
\resizebox{1.9\columnwidth}{!}{
\begin{tabular}{lccc|ccc}
\toprule
Method & Struct. Const. & RAG & Self-Corr. & Precision & Recall & F1 \\
\midrule
\citet{Li2025_LLMsDoWell_SRL} (LLaMA-8B) & \xmark & \cmark & \cmark & 6.29 & 7.59 & 6.88 \\
Ours (gpt-oss-120b) & \cmark & \xmark & \xmark & 46.58 & 35.58 & 42.17 \\
Ours (gpt-oss-120b with Dependency-Guided Structure) & \cmark & \xmark & \xmark & \textbf{49.03} & \textbf{40.44} & \textbf{44.30} \\
\bottomrule
\end{tabular}
}
\caption{Comparison of frozen on SRL performance across inference strategies on the CoNLL 2012 test set. Struct. Const.(Structural Constraints) denotes schema-constrained prompts enforcing BIO tagging rules.}
\label{tab:LLM}
\vspace*{-.15in}
\end{table*}

\section{Representation Behavior of LLMs}

Building on the structured SRL framework, we investigate how LLMs perform SRL when provided explicit syntactic structural. We examine whether externally supplied dependency information influences SRL predictions and how reliably LLMs utilize such signals. To enable controlled analysis, we compare LLM outputs with and without dependency-informed context, isolating the interaction between syntactic structure and generative semantic inference. Experiments are conducted in a zero-shot setting using gpt-oss-120b.

To probe LLM SRL behavior under controlled conditions, we adopt an instructional guidance formulation strategy,\footnote{See Appendix~\ref{appendix_Prompt} for the full prompt.} providing explicit role definitions and structural output constraints on the CoNLL-2012 test set. We compare predictions with and without dependency-informed syntactic signals, enabling a controlled analysis of how external structural signals influences SRL performance.

Table~\ref{tab:LLM} shows that dependency-informed prompts improve predictions (F1: 42.17 vs. 44.30). The gains reflect improved span-boundary stability and phrase-level coherence rather than role classification accuracy, indicating more reliable argument grouping. Unlike prior LLM-based SRL approaches requiring retrieval or iterative self-correction~\cite{Li2025_LLMsDoWell_SRL}, our formulation remains lightweight, relying on structural guidance and dependency cues.

Qualitative analysis further illustrates these structural effects. Consider \textit{Fruit that shows a hint of black through the red color of the skin fetches the best price}, where \textit{fetches} is the predicate. Without dependency cues, LLMs collapse argument spans, assigning ARG0 only \textit{Fruit}. With dependency information, the model preserves the full phrase (\textit{Fruit that shows a hint of black through the red color of the skin}), indicating improved structural grouping. 

A second pattern involves duplicate role assignments. In \textit{to publicly support the project}, predictions without dependency signals produce fragmented ARG1 spans \textit{[B-ARG1: to I-ARG1: publicly] [B-ARG1: support I-ARG1: the I-ARG1: project]}. Dependency-informed prompts instead yield a single continuous span, reflecting constraint-consistent phrase composition.

These findings indicate that dependency-informed structural signals enhance representation stability rather than classification capacity, highlighting the interaction between syntactic structure and generative semantic inference.

\section{Multilingual Applicability: A Projection Case Study} 

%SRL models %are 
%have been predominantly developed for English 
%due to the scarcity of large, 
%because they rely on large amounts of 
%human-labeled resources in other languages.%
%data. 
%However, such annotated resources are scarce for many other languages, and w
%Without sufficient training data, it is difficult to build high-quality SRL models. 
Building on our English benchmark results, we extend the model to multilingual transfer settings. 
% \bonnieshort{I don't want to rehash background about scarcity of data here, but get right to the point. Edited a bit.}
%To address this limitation, 
We adopt cross-lingual SRL projection techniques to examine how explicitly represented predicate-argument structures transfer across-languages, aligning English–French sentence pairs via one-to-one token mappings prior to semantic role transfer.

% generate multilingual datasets that support SRL development across different languages. 

Recent work on AllenNLP-based cross-lingual SRL transfer highlights hallucinations arising from structural mismatches between languages~\cite{youm2024dahrs}.
%caused by cross-linguistic differences 
%and adapting to linguistic distinctions (e.g., head-final vs head-initial)
%when transferring semantic 
The approach aligns English-French sentence pairs by first adding or removing tokens to ensure one-to-one mappings prior to semantic role transfer. Adding or removing tokens to ensure one-to-one mappings prior to SRL transfer still carry over errors already present in the English SRL outputs. 

Figure~\ref{tab:dahrs} illustrates such a case. When aligning \textit{after the October 1987 crash} with its French counterpart \textit{apr\`{e}s l'\'{e}crasement (de) octobre 1987}, \textit{de} is omitted and \textit{\'{e}crasement} is repeated. However, on the English side, \textit{after} is incorrectly tagged as \texttt{I-ARGM-TMP} rather than \texttt{B-ARGM-TMP}, leaving the correction incomplete and propagating the error to the French side.

% \textbf{[put in shortcomings here with real examples, including ones we discussed, including BIO issues.]}\bonnieshort{Take note and edit.}
\begin{figure}[h!]
\scriptsize{
\begin{flushleft}
% \textbf{Before FCFA}\\
\begin{tabular}{ l c l }
% \textcolor{purple}{[I-ARG1] circuit} & --- & \textcolor{blue}{$\epsilon$} \\ 
% \textcolor{purple}{[I-ARG1] breakers} & --- & \textcolor{blue}{disjoncteurs}\textcolor{black}{; 4-2}\\
% \textcolor{purple}{[B-V] installed} & --- & \textcolor{blue}{install\'es}\textcolor{black}{; 6-4}\\
\textcolor{purple}{[I-ARGM-TMP] after} & --- & \textcolor{blue}{apr\`es}\textcolor{black}{; 7-5}\\
\textcolor{purple}{[I-ARGM-TMP] the} & --- & \textcolor{blue}{l'} \textcolor{black}{; 8-6}\\ 
\textcolor{purple}{[I-ARGM-TMP] october} & --- & \textcolor{blue}{écrasement} \textcolor{black}{; 9-7}\\ 
\textcolor{purple}{[I-ARGM-TMP] october} & --- & \textcolor{blue}{octobre} \textcolor{black}{; 9-9}\\ 
\textcolor{purple}{[I-ARGM-TMP] 1987} & --- & \textcolor{blue}{1987} \textcolor{black}{; 10-10}\\ 
\textcolor{purple}{[I-ARGM-TMP] crash} & --- & \textcolor{blue}{écrasement} \textcolor{black}{; 11-7}
\end{tabular}
\end{flushleft}
}
\vspace*{-.1in}
\caption{Prior cross-lingual SRL~\cite{youm2024dahrs} resolves alignment errors before SRL transfer. The English excerpt (tokens 7-11) aligns with the French excerpt (tokens 5-10). English tokens \textit{october} (9) and \textit{crash} (11) align to French \textit{\'ecrasement} (7). The spurious 9-7 alignment is ultimately excluded, but \textit{after}---and correspondingly \textit{apr\`{e}s}---are left incorrectly tagged as \texttt{I-ARG-TMP}.}
\label{tab:dahrs}
\vspace*{-.1in}
\end{figure}

%In the transfer process, the AllenNLP SRL labeler is commonly used; however, 
We address this limitation by replacing the AllenNLP SRL component with our structure inference framework, preserving predicate-argument boundaries while eliminating redundant per-predicate processing. This restructuring yields $\approx 10\times$ faster projection and enables large-scale multilingual dataset construction.

% \bonnieshort{is it 9 times or 10 times? I've seen both numbers floating around?}\sangpilshort[]{it is 10 times}

Beyond efficiency, our model improves structural accuracy. In the same example, the dependency-aware detector recognizes \textit{after the October 1987 crash} as a single noun phrase and labels it under \texttt{pp\_attach}. It then correctly initiates the span with a \texttt{B-ARGM-TMP} tag on \textit{after}, rather than the erroneous \texttt{I-ARGM-TMP} label in Figure~\ref{tab:dahrs-fixed}.
% \bonnieshort{These aren't tables, they are figures.}

\begin{figure}[h!]
\begin{flushleft}

\scriptsize
\begin{tabular}{ l c l }

% \textcolor{purple}{[I-ARG1] circuit} & --- & \textcolor{blue}{$\epsilon$} \\ 
% \textcolor{purple}{[I-ARG1] breakers} & --- & \textcolor{blue}{disjoncteurs}\textcolor{black}{}\\
% \textcolor{purple}{[B-V] installed} & --- & \textcolor{blue}{install\'es}\textcolor{black}{}\\
\textcolor{purple}{[B-ARGM-TMP] after} & --- & \textcolor{blue}{[B-ARGM-TMP] apr\`es}\textcolor{black}{}\\
\textcolor{purple}{[I-ARGM-TMP] the} & --- & \textcolor{blue}{[I-ARGM-TMP] l'} \textcolor{black}{}\\ 
% \textcolor{purple}{[I-ARGM-TMP] october} & --- & \textcolor{blue}{écrasement} \textcolor{black}{}\\ 
\textcolor{purple}{[I-ARGM-TMP] october} & --- & \textcolor{blue}{[I-ARGM-TMP] octobre} \textcolor{black}{}\\ 
\textcolor{purple}{[I-ARGM-TMP] 1987} & --- & \textcolor{blue}{[I-ARGM-TMP] 1987} \textcolor{black}{}\\ 
\textcolor{purple}{[I-ARGM-TMP] crash} & --- & \textcolor{blue}{[I-ARGM-TMP] écrasement} \textcolor{black}{}
\end{tabular}
\end{flushleft}
\vspace*{-.1in}
\caption{Dependency-aware correction applied to the \textit{October 1987 crash} example. The revised SRL output correctly marks after with a \texttt{B-ARGM-TMP} tag, restoring the missing boundary and producing a complete span alignment across English and French.}
\label{tab:dahrs-fixed}
\vspace*{-.15in}
\end{figure}
% \vspace*{-.15in}

These findings underscore a broader implication: structurally explicit semantic representations provide stability advantages not only within monolingual inference but also during cross-lingual projection, where small boundary deviations can amplify into large inconsistencies. Leveraging this structure-preserving design, our framework supports structurally faithful SRL projection in downstream multilingual settings; while illustrated here through an English-French case study, this analysis highlights the potential value of structured SRL for multilingual applications that leverage English-centric resources.

% Because our model is lightweight, up-to-date, and fully integrated from preprocessing to training, it forms a foundation for future multilingual SRL systems that balance speed, scalability, and structural fidelity.
%it is well-positioned to serve as a foundation for future SRL systems.

\section{Conclusion and Future Work}

This paper revisits structured semantic role labeling (SRL) through a modernized encoder-based framework that eliminates redundant per-predicate encoding while preserving predictive performance. By caching sentence-level representations, the framework achieves a 10$\times$ inference speedup over AllenNLP while maintaining comparable accuracy, and supports stronger performance with alternative encoders such as RoBERTa and DeBERTa.

We also introduce a dependency-informed diagnostic methodology for systematically characterizing span boundary inconsistencies. The analysis shows that dependency-informed structural signals primarily enhance representational stability in both encoder-based and LLM-based SRL, improving span coherence and argument grouping rather than raw classification accuracy. The framework aligns naturally with cross-lingual SRL projection pipelines, with implications for downstream multilingual SRL transfer.

Future work includes fine-tuning decoder-only LLMs within this structured formulation, integrating SRL transfer pipelines for multilingual dataset construction, and refining dependency-aware error detection to further reduce regression cases. Limitations of the framework and ethical considerations are discussed in Appendices~\ref{app:limitation} and~\ref{app:ethical}.\footnote{See Appendix~\ref{app:limitation} for limitations and Appendix~\ref{app:ethical}  for ethical considerations.}

\section*{Acknowledgement}

This research is based upon work supported, in part, by the Defense Advanced Research Projects Agency (DARPA) under Contract No. W912CG-24-C-0014 and, in part, by the National Institute Of Mental Health of the National Institutes of Health under Award Number R01MH135504. The views and conclusions expressed in this paper are those of the authors, who are solely responsible for its content, and do not necessarily represent the official views, policies, or positions, either expressed or implied, of the National Institutes of Health, DARPA, or the U.S. Government. The U.S. Government is authorized to reproduce and distribute reprints for Government purposes notwithstanding any copyright notation hereon.

\bibliography{bibliography}

\appendix

\section{Dataset Creation}
\label{sec:app_dataset}

In CoNLL-2012, each predicate in a sentence yields a separate SRL instance, resulting in multiple entries for sentences containing more than one predicate. As shown in Figure~\ref{fig:json-object-output}, each instance is stored as a JSON object containing:

\begin{center}
\footnotesize{
\begin{itemize}
\item \texttt{words}: the tokenized sentence;
\item \texttt{predicate\_word\_idx}: the index of the predicate token; and
\item BIO-formatted sequence of semantic role tags (e.g., \texttt{["B-ARG0", "B-V", "B-ARG2", "B-ARG1", "I-ARG1", "O"]}).
\end{itemize}
}
\end{center}

\begin{figure}
\footnotesize{
\begin{tcolorbox}[
    % title=Example JSON Data,
    fonttitle=\bfseries\sffamily\small, % Customize title font
    colback=white, % Background color
    boxsep=1mm,   % Spacing between text and box border
    left=1mm,     % Indentation for text
    right=1mm,
    top=1mm,
    bottom=1mm,
    boxrule=0.5pt, % Border line thickness
    sharp corners, % Use sharp, not rounded, corners
    halign=left,  % Left-align the title
]
\ttfamily\footnotesize % Monospace, small font for the content
{\{

  'words': ['Mary', 'gave', 'me', 'a', 'present', '.'], \\
  'predicate\_word\_idx': 1, \\
  'labels': ['B-ARG0', 'B-V', 'B-ARG2', 'B-ARG1', 'I-ARG1', 'O']
  
  \}
}
\end{tcolorbox}
}
% \vspace*{-.17in}
\caption{Example SRL instance represented as a JSON object, showing tokenized words, the predicate index, and BIO-formatted semantic role labels.}
\label{fig:json-object-output}
\vspace*{-.2in}
\end{figure}

To mirror AllenNLP's data format, we reconstruct tokenized \texttt{.gold\_skel} files from OntoNotes \texttt{.onf} files as follows:

\begin{center}
\footnotesize{
\begin{itemize}
\item \textbf{Token extraction:} Extract tokens and their indices from each 
    %The tokens for each sentence and their corresponding indices are extracted from the 
    parse tree, removing 
    %for each section in the \texttt{.onf} files. Tokens containing 
    artifacts such as ``$\ast$'' or ``$\%$.''
\item \textbf{Gold skeleton alignment:} Align the extracted tokens with corresponding
    %Corresponding formatted 
    \texttt{.gold\_skel} sections,
    replacing 
    %are paired with the extracted tokens, replacing the
    \texttt{[WORD]} placeholders
    %tokens with the aligned word sequence.
    with surface tokens.
\item \textbf{Semantic role reconstruction:} Convert column-based % Column-based 
    SRL annotations (columns $\geq 12$ in CoNLL format) 
    %are parsed from each section. The bracketed span annotations are converted 
    into BIO tags 
    %(e.g., \texttt{(ARG1*)} $\rightarrow$ \texttt{B-ARG1}) 
    using a custom parser that handles nested and single-token spans.
\item \textbf{JSON object generation:} For each predicate, build
    %the script builds 
    a JSON object capturing the predicate index, argument role spans (\texttt{ARG0}, \texttt{ARG1}, \texttt{ARG2}, etc.), and the complete BIO label sequence. Sentences with misaligned tokens are
    %with inconsistent token–role alignment are 
    logged and excluded.
\end{itemize}
}
\end{center}

This preprocessing yields a large, sentence-aligned set of predicate-level instances %used 
for fine-tuning and evaluation, fully compatible with prior SRL systems. Each example represents %corresponds to 
a specific predicate and its arguments.

\section{Distribution of Missing Semantic Roles\textemdash Our system}
\label{sec:missing_roles_our}
\begin{figure}[htbp]
    \centering
    \includegraphics[width=.8\linewidth]{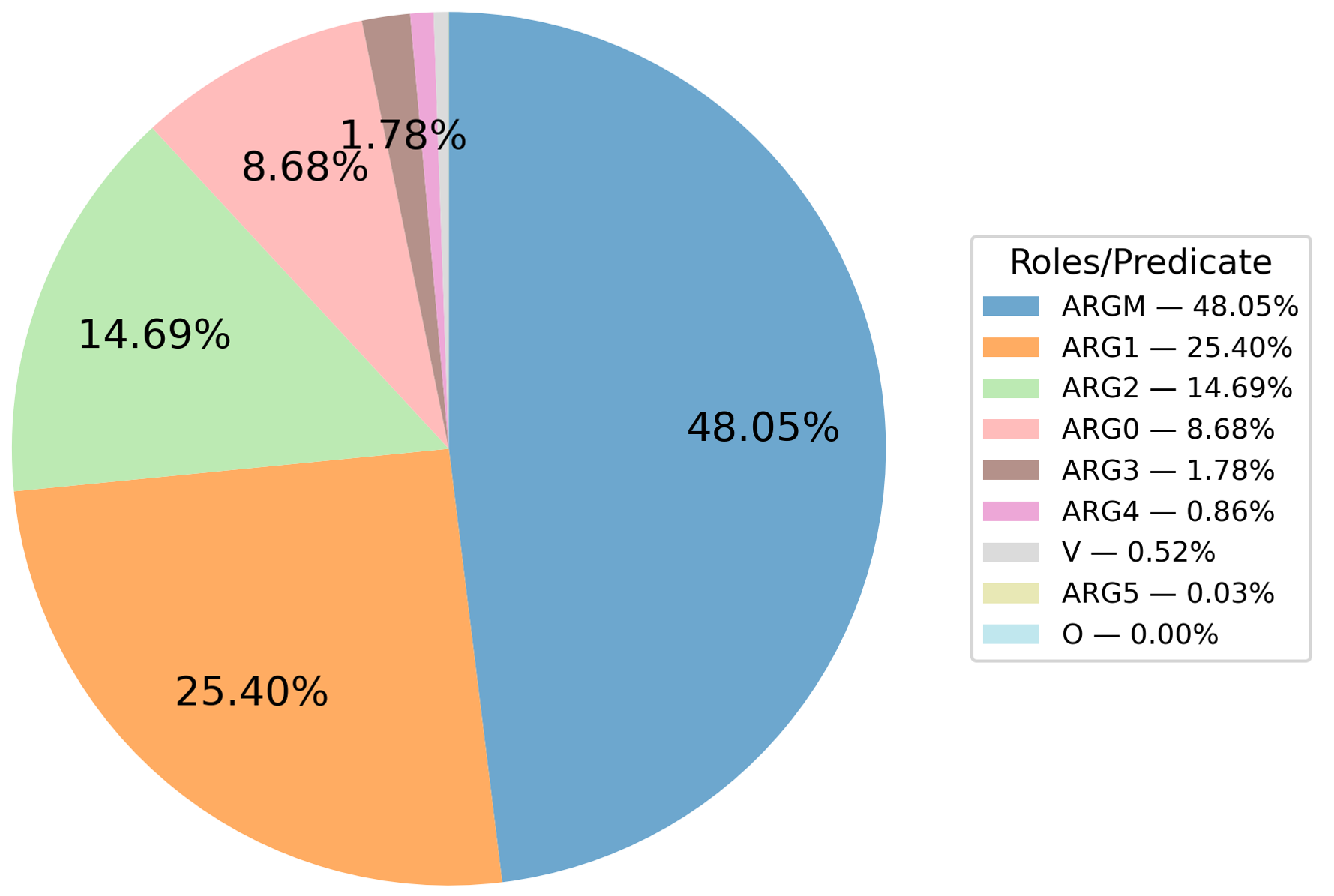}
    
    \caption{Proportion of missing roles or predicates in \textbf{our system's} predictions compared to OntoNotes 5.0 ground truth. }%Process of sublingual translations and LLM prompting
    % \vspace{-.1in}
    \label{fig:diff_ours_args}
    \vspace*{-.1in}
\end{figure}

\begin{figure}[htbp]
    \centering
    \includegraphics[width=.8\linewidth]{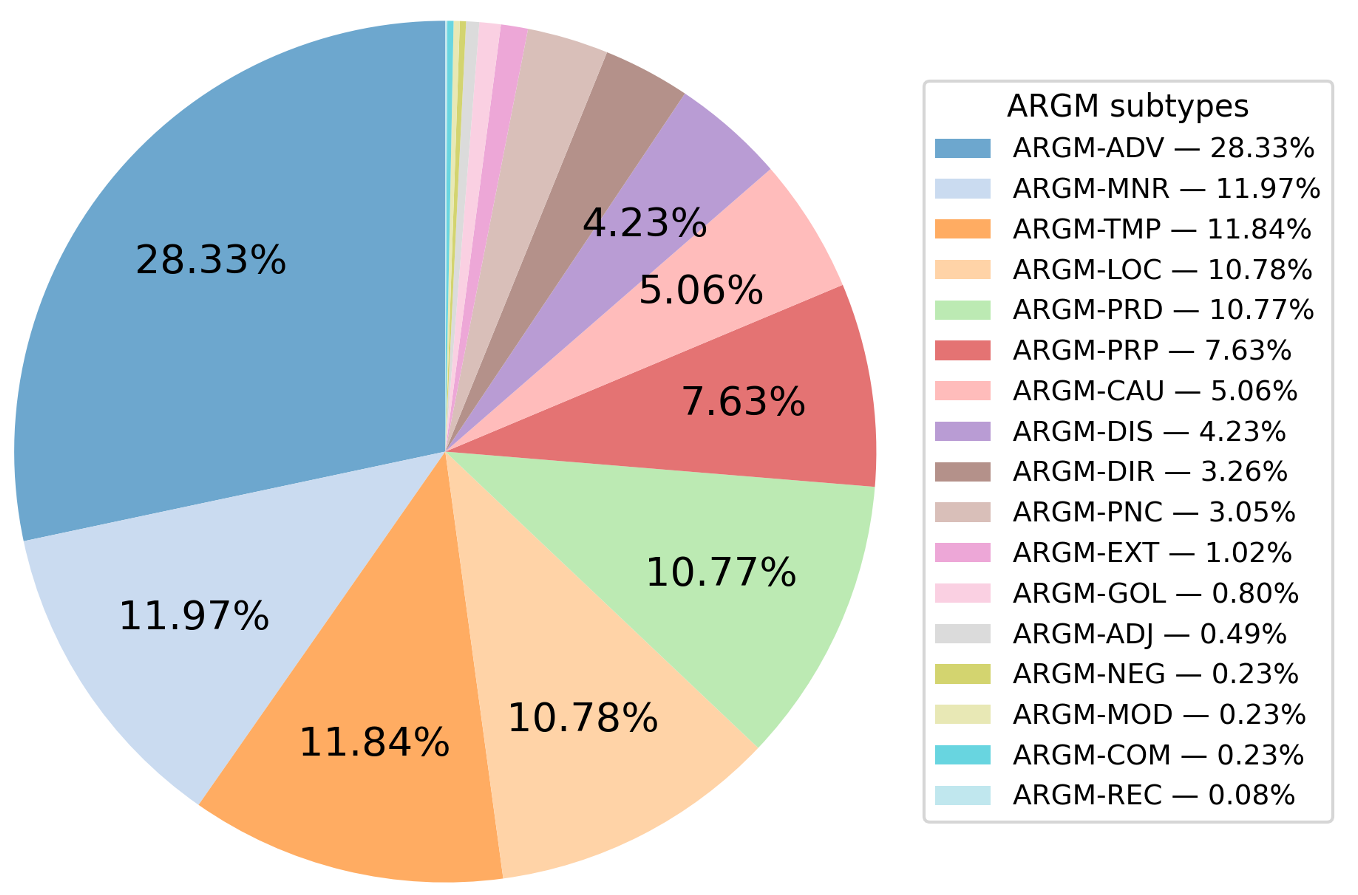}
    
    \caption{Proportion of missing ARGMs in \textbf{our system's} predictions compared to OntoNotes 5.0 ground truth.
    }%Process of sublingual translations and LLM prompting
    % \vspace{-.1in}
    \label{fig:diff_argms_oursystem}
    \vspace*{-.1in}
\end{figure}

This section summarizes the distribution of missing roles in \textbf{our system} relative to the ground truth. Figure~\ref{fig:diff_ours_args} shows the proportions of roles missing from our predictions, while Figure~\ref{fig:diff_argms_oursystem} presents the proportions of missing modifiers (ARGMs) from our prediction.

\section{Distribution of Missing Semantic Roles\textemdash AllenNLP}
\label{sec:missing_roles_allen}
\begin{figure}[htbp]
    \centering
    \includegraphics[width=.8\linewidth]{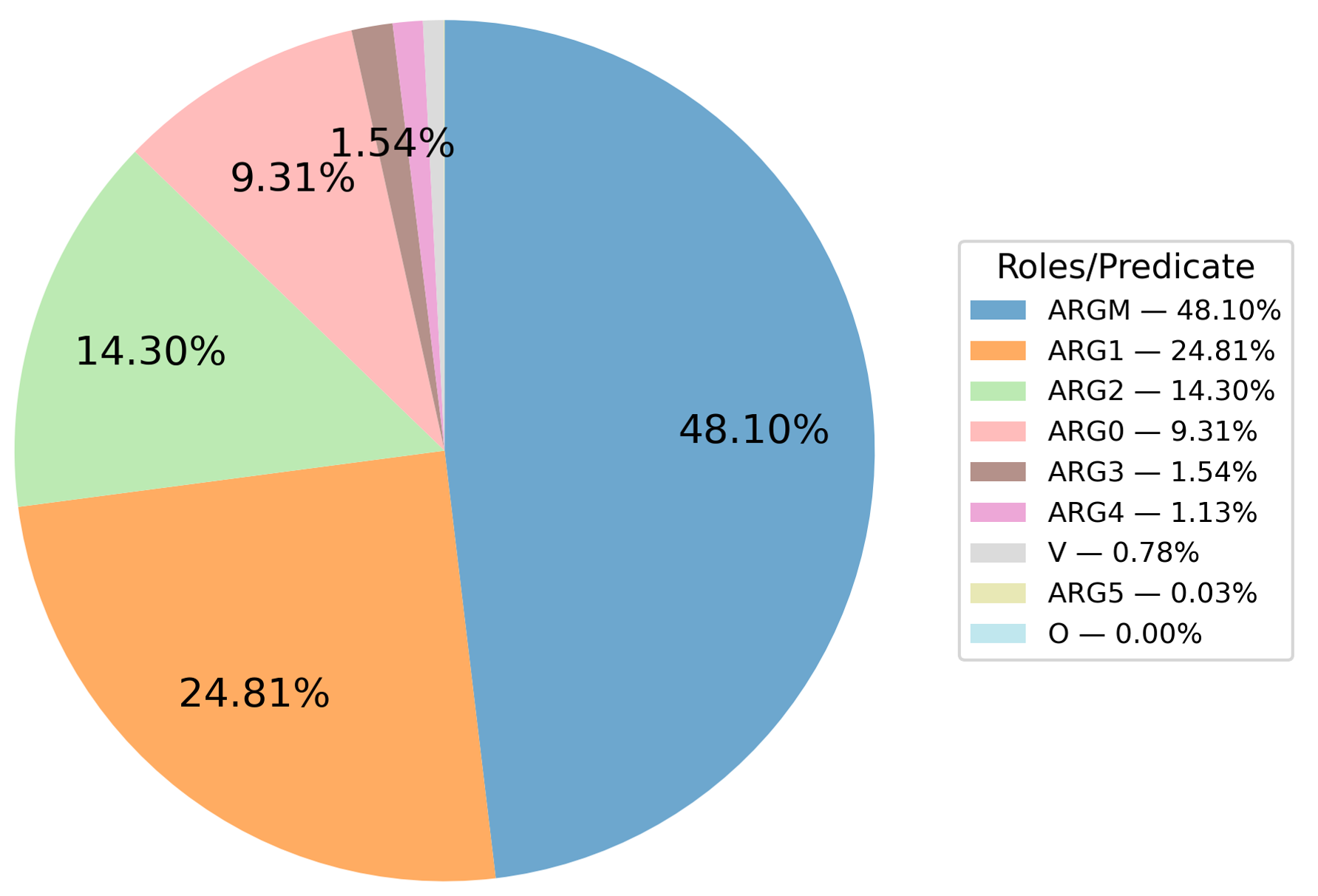}
    
    \caption{Proportion of missing roles or predicates in \textbf{AllenNLP's} predictions compared to OntoNotes 5.0 ground truth.
    }%Process of sublingual translations and LLM prompting
    % \vspace{-.1in}
    \label{fig:diff_allen_argMs}
    \vspace*{-.1in}
\end{figure}

\begin{figure}[htbp]
    \centering
    \includegraphics[width=.8\linewidth]{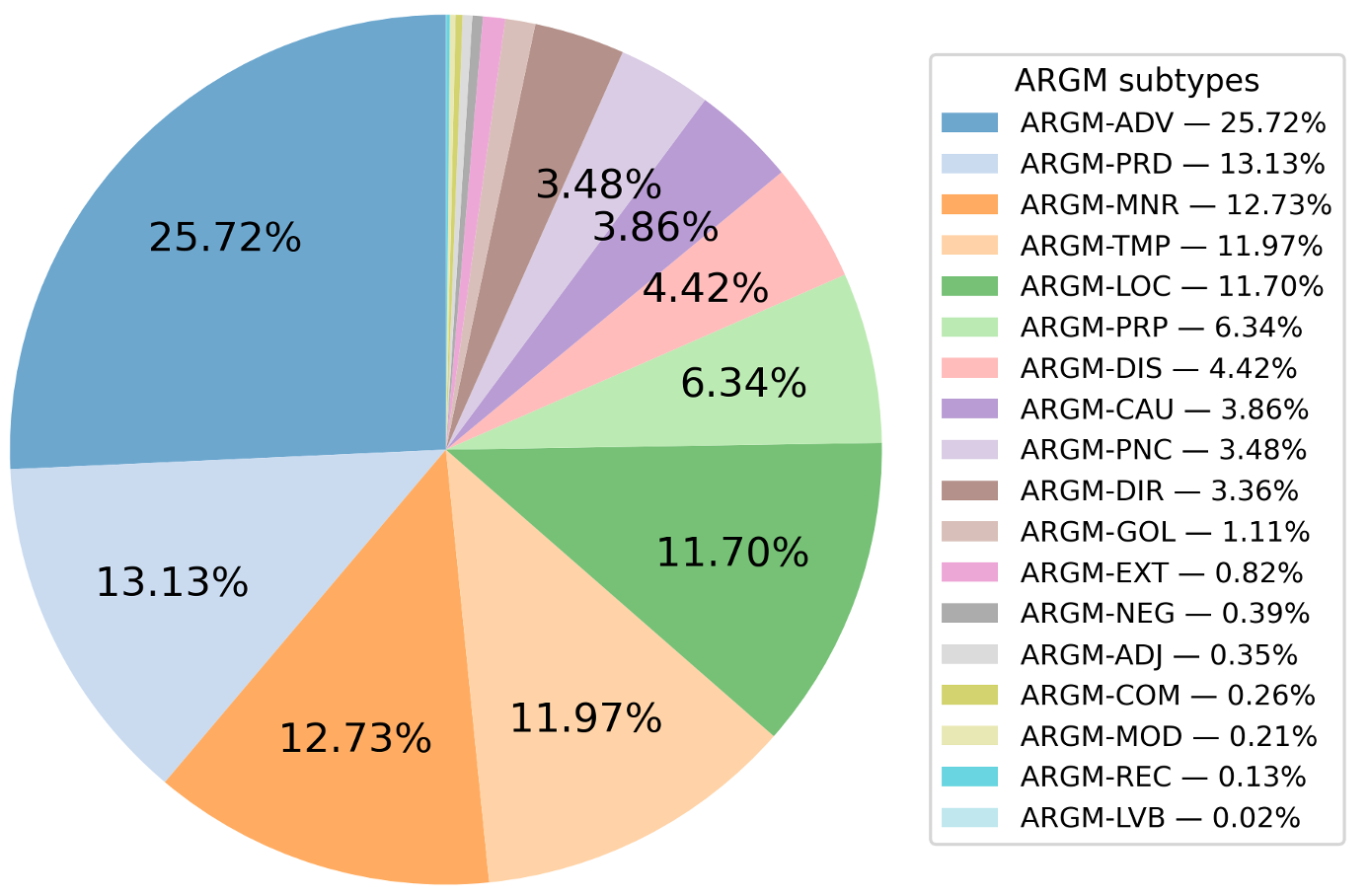}
    
    \caption{Proportion of missing ARGMs in \textbf{AllenNLP's} predictions compared to OntoNotes 5.0 ground truth.
    }%Process of sublingual translations and LLM prompting
    % \vspace{-.1in}
    \label{fig:diff_argms_allen}
    \vspace*{-.1in}
\end{figure}

This section summarizes the distribution of missing roles in \textbf{AllenNLP} relative to the ground truth. Figure~\ref{fig:diff_allen_argMs} reports the proportions of roles missing from AllenNLP, while Figure~\ref{fig:diff_argms_allen} shows the proportions of missing modifiers (ARGMs) from AllenNLP prediction.

\section{Error Analyzer Overview}
\label{appendix:error}

\begin{figure}[htbp]
    \centering
    \includegraphics[width=0.7\linewidth]{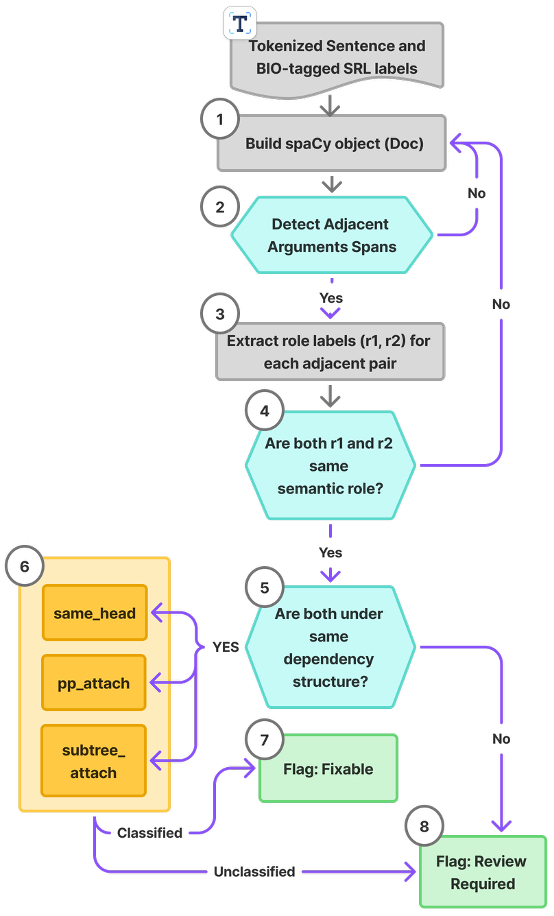}
    
    \caption{Overview of the modernized Dependency Aware Error Analyzer.
    }%Process of sublingual translations and LLM prompting
    % \vspace{-.1in}
    \label{fig:error}
    \vspace*{-.1in}
\end{figure}

Figure~\ref{fig:error} presents the dependency-aware error analysis pipeline, detailing the sequential steps used to detect adjacent spans, assess semantic role consistency, and categorize structural inconsistencies.

\section{Example Prompts for LLM Evaluation}
\label{appendix_Prompt}
This section presents the prompt used for dependency-informed SRL, shown both \textbf{with} and \textbf{without} dependency information in the LLM.

\begin{figure*}[t]
\footnotesize
\centering
\begin{tcolorbox}[
  title={Prompt with dependency information for SRL},
  colback=gray!5,
  colframe=black!60,
  fonttitle=\bfseries,
  enhanced,
  % breakable,
  width=\textwidth,
  left=1.2mm,right=1.2mm,top=1mm,bottom=1mm
]
\footnotesize
\setlength{\parindent}{0pt}

\textbf{sent (MUST NOT change):} I want to go home\\

\smallskip

\textbf{Phrase hints (token spans over Words):}\\
NP0: NP [0:1], head=0, text="I"\\
VP3: VP [2:5], head=3, text="to go home"\\
XCOMP1: XCOMP [2:5], head=3, text="to go home"\\
VP2: VP [0:5], head=1, text="I want to go home"

\smallskip
\textbf{Protected phrase IDs (DO NOT split SRL arguments inside these spans):} ["XCOMP1"]

\smallskip
\textbf{Return SRL in EXACTLY this JSON schema (example):}
\begin{lstlisting}[basicstyle=\ttfamily\footnotesize, columns=fullflexible, breaklines=true]
{
  "sent": "",
  "words": ["..."],
  "verbs": [
    {"verb": "...", "label": ["O","B-ARG0","I-ARG0","B-V"]}
  ]
}
\end{lstlisting}

\smallskip
\textbf{Hard rules:}\\
(1) words is the list of tokens obtained from the sentence, separated by whitespace \quad
(2) For each verb, label length MUST equal len(words). \quad
(3) BIO: I-X cannot start a span; it must follow B-X/I-X of the same X. \quad
(4) Include ALL predicates. \quad
(5) Use phrase hints ONLY as boundary guidance; do NOT create new tokens.\quad
(6) Phrase integrity: do NOT split a role into multiple segments inside any protected phrase span.\quad
\textbf{Bad:} [ARG1: to] [ARG1: go home] when 'to go home' is protected.

\smallskip
\textbf{Label guide:}\\
\textbf{Core:} ARG0 agent; ARG1 patient; ARG2 instrument/benefactive/attribute; ARG3 start/benefactive/attribute; ARG4 end; ARG5 predicate-dependent.\\
\textbf{Modifiers:} ARGM-EXT extent; ARGM-LOC location; ARGM-DIR direction; ARGM-NEG negation; ARGM-MOD general-mod; ARGM-ADV adverbial; ARGM-MNR manner;\\
ARGM-PRD secondary-pred; ARGM-REC reciprocal; ARGM-TMP temporal; ARGM-PRP purpose; ARGM-PNC purpose-no-cause; ARGM-CAU cause; ARGM-ADJ adjectival (nouns);\\
ARGM-COM comitative; ARGM-DIS discourse; ARGM-DSP direct-speech; ARGM-GOL goal; ARGM-LVB light-verb (nouns); ARGA secondary-agent; ARGM-PRR predicating-relation.

\end{tcolorbox}
\end{figure*}

\begin{figure*}[t]
\footnotesize
\centering
\begin{tcolorbox}[
  title={Prompt without dependency information for SRL},
  colback=gray!5,
  colframe=black!60,
  fonttitle=\bfseries,
  enhanced,
  % breakable,
  width=\textwidth,
  left=1.2mm,right=1.2mm,top=1mm,bottom=1mm
]
\footnotesize
\setlength{\parindent}{0pt}

\textbf{sent (MUST NOT change):} I want to go home\\

\smallskip

\textbf{Return SRL in EXACTLY this JSON schema (example):}
\begin{lstlisting}[basicstyle=\ttfamily\footnotesize, columns=fullflexible, breaklines=true]
{
  "sent": "",
  "words": ["..."],
  "verbs": [
    {"verb": "...", "label": ["O","B-ARG0","I-ARG0","B-V"]}
  ]
}
\end{lstlisting}

\smallskip
\textbf{Hard rules:}\\
(1) words is the list of tokens obtained from the sentence, separated by whitespace \quad
(2) For each verb, label length MUST equal len(words). \quad
(3) BIO: I-X cannot start a span; it must follow B-X/I-X of the same X.\quad
(4) Include ALL predicates in the sentence.

\smallskip
\textbf{Label guide:}\\
\textbf{Core:} ARG0 agent; ARG1 patient; ARG2 instrument/benefactive/attribute; ARG3 start/benefactive/attribute; ARG4 end; ARG5 predicate-dependent.\\
\textbf{Modifiers:} ARGM-EXT extent; ARGM-LOC location; ARGM-DIR direction; ARGM-NEG negation; ARGM-MOD general-mod; ARGM-ADV adverbial; ARGM-MNR manner;\\
ARGM-PRD secondary-pred; ARGM-REC reciprocal; ARGM-TMP temporal; ARGM-PRP purpose; ARGM-PNC purpose-no-cause; ARGM-CAU cause; ARGM-ADJ adjectival (nouns);\\
ARGM-COM comitative; ARGM-DIS discourse; ARGM-DSP direct-speech; ARGM-GOL goal; ARGM-LVB light-verb (nouns); ARGA secondary-agent; ARGM-PRR predicating-relation.

\end{tcolorbox}
\end{figure*}

\section{Limitations}
\label{app:limitation}

We analyze the errors produced by our SRL models and categorize them using a dependency-aware detector. While the detector identifies systematic split-span patterns, the majority of cases still require human judgment for comprehensive resolution. To mitigate this, we report which errors are automatically solvable according to our rules and which are not safely fixable without review. This yields an actionable map of error types, clarifies the limits of automatic repair, and guides future SRL diagnostics by focusing human effort where it is most impactful.

\section{Ethics Statement on Broader Impact}
\label{app:ethical}

Our SRL model relies on Transformer backbones trained on OntoNotes, which may encode the domain, genre, and annotators' narrow view present in that corpus. As a result, the model can systematically underperform\textemdash or ``penalize''\textemdash certain argument types or constructions that are underrepresented in OntoNotes, and these errors are not always fixable by automatic post-hoc analysis. To mitigate this, we encourage training and evaluation on broader, contemporary corpora, and we release rules/code to enable independent audits. Through this process, future work aims to narrow gaps rather than overfitting to OntoNotes.

\end{document}